\documentclass[a4paper,twoside]{article}

\usepackage[dvipsnames,svgnames,x11names]{xcolor}
\usepackage{tikz}
\usetikzlibrary{arrows.meta,shapes,calc,matrix,fit,positioning,backgrounds,decorations.markings,fadings}
\usepackage{pgfplots}
\usepackage{pgfplotstable}
\pgfplotsset{compat=1.9}
\usepackage{xstring}

\usepgfplotslibrary{external}

\IfBeginWith*{\jobname}{fig/extern/}{\finalcopy}{}

\tikzstyle{every picture}+=[
	remember picture,
	every text node part/.style={align=center},
	every matrix/.append style={ampersand replacement=\&},
]
\tikzstyle{tight} = [inner sep=0pt,outer sep=0pt]
\tikzstyle{node}  = [draw,circle,tight,minimum size=12pt,anchor=center]
\tikzstyle{op}    = [draw,circle,tight]
\tikzstyle{dot}   = [fill,draw,circle,inner sep=1pt,outer sep=0]
\tikzstyle{pt}    = [fill,draw,circle,inner sep=1.5pt,outer sep=.2pt]
\tikzstyle{box}   = [draw,rectangle,inner sep=3pt]
\tikzstyle{high}  = [black!60]
\tikzstyle{group} = [high,box,opacity=.5]
\tikzstyle{dim1}  = [fill opacity=.3,text opacity=1]
\tikzstyle{dim2}  = [fill opacity=.5,text opacity=1]
\tikzstyle{dim3}  = [fill opacity=.7,text opacity=1]
\tikzstyle{rectc} = [tight,transform shape]
\tikzstyle{rect}  = [rectc,anchor=south west]

\tikzset{every mark/.append style={solid}}
\pgfplotsset{
	grid=both, width=\columnwidth, try min ticks=5,
	every axis/.append style={font=\small},
	every axis plot/.append style={thick,mark=none,mark size=1.8,tension=0.18},
	legend cell align=left, legend style={fill opacity=0.8},
	xticklabel={\pgfmathprintnumber[assume math mode=true]{\tick}},
	yticklabel={\pgfmathprintnumber[assume math mode=true]{\tick}},
	nodes near coords math/.style={
		nodes near coords={\pgfmathprintnumber[assume math mode=true]{\pgfplotspointmeta}},
	},
}

\pgfplotsset{
	dash/.style={mark=o,dashed,opacity=0.6},
	dott/.style={mark=o,dotted,opacity=0.6},
	nolim/.style={enlargelimits=false},
	plain/.style={every axis plot/.append style={},nolim,grid=none},
}

\pgfdeclarelayer{bg4}
\pgfdeclarelayer{bg3}
\pgfdeclarelayer{bg2}
\pgfdeclarelayer{bg1}
\pgfdeclarelayer{fg1}
\pgfdeclarelayer{fg2}
\pgfdeclarelayer{fg3}
\pgfdeclarelayer{fg4}
\pgfsetlayers{bg4,bg3,bg2,bg1,main,fg1,fg2,fg3,fg4}

\pgfkeys{/tikz/.cd, aspect/.store in=\aspect, aspect=1}
\pgfkeys{/tikz/.cd, depth/.store in=\depth, depth=.5}
\pgfkeys{/tikz/.cd, stepx/.store in=\step, stepx=1}

\tikzstyle{geom} = [line join=bevel,aspect=1,depth=.5,z={(\depth*\aspect,\depth)}]
\tikzstyle{wire} = [geom,draw,thick]

\def\cx[#1,#2,#3]{#1}
\def\cy[#1,#2,#3]{#2}
\def\cz[#1,#2,#3]{#3}
\def\ex[#1,#2,#3]{#1,0,0}
\def\ey[#1,#2,#3]{0,#2,0}
\def\ez[#1,#2,#3]{0,0,#3}

\makeatletter
\renewcommand\paragraph{\@startsection{paragraph}{4}{\z@}{1ex}{-1em}{\normalfont\normalsize\bfseries}}
\makeatother

\usepackage{times}
\usepackage{apalike}

\usepackage{epsfig}
\usepackage{subcaption}
\usepackage{calc}
\usepackage{amsfonts}
\usepackage{bbm}
\usepackage{amssymb}
\usepackage{amstext}
\usepackage{amsmath}
\usepackage{amsthm}
\usepackage{multicol}
\usepackage[ruled,vlined]{algorithm2e}
\usepackage[bottom]{footmisc}
\usepackage{enumitem}
\usepackage{booktabs}
\usepackage{float}
\usepackage{multirow}
\usepackage{algorithmicx}
\usepackage[pagebackref,breaklinks,colorlinks]{hyperref}
\usepackage[capitalize]{cleveref}
\floatstyle{plaintop}
\restylefloat{table}

\usepackage{SCITEPRESS}     

\begin{document}

\newcommand{\head}[1]{{\smallskip\noindent\textbf{#1}}}
\newcommand{\alert}[1]{{\color{red}{#1}}}
\newcommand{\sm}{\scriptsize}
\newcommand{\eq}[1]{(\ref{eq:#1})}

\newcommand{\Th}[1]{\textsc{#1}}
\newcommand{\mr}[2]{\multirow{#1}{*}{#2}}
\newcommand{\mc}[2]{\multicolumn{#1}{c}{#2}}
\newcommand{\tb}[1]{\textbf{#1}}
\newcommand{\ch}{\checkmark}

\newcommand{\red}[1]{{\color{red}{#1}}}
\newcommand{\blue}[1]{{\color{blue}{#1}}}
\newcommand{\green}[1]{{\color{green}{#1}}}
\newcommand{\gray}[1]{{\color{gray}{#1}}}

\newcommand{\citeme}[1]{\red{[XX]}}
\newcommand{\refme}[1]{\red{(XX)}}

\newcommand{\fig}[2][1]{\includegraphics[width=#1\linewidth]{fig/#2}}
\newcommand{\figh}[2][1]{\includegraphics[height=#1\linewidth]{fig/#2}}

\newcommand{\tran}{^\top}
\newcommand{\mtran}{^{-\top}}
\newcommand{\zcol}{\mathbf{0}}
\newcommand{\zrow}{\zcol\tran}

\newcommand{\ind}{\mathbbm{1}}
\newcommand{\expect}{\mathbb{E}}
\newcommand{\nat}{\mathbb{N}}
\newcommand{\zahl}{\mathbb{Z}}
\newcommand{\real}{\mathbb{R}}
\newcommand{\proj}{\mathbb{P}}
\newcommand{\prob}{\operatorname{P}}
\newcommand{\normal}{\mathcal{N}}

\newcommand{\mif}{\textrm{if}\ }
\newcommand{\other}{\textrm{otherwise}}
\newcommand{\minimize}{\textrm{minimize}\ }
\newcommand{\maximize}{\textrm{maximize}\ }
\newcommand{\st}{\textrm{subject\ to}\ }

\newcommand{\id}{\operatorname{id}}
\newcommand{\const}{\operatorname{const}}
\newcommand{\sgn}{\operatorname{sgn}}
\newcommand{\var}{\operatorname{Var}}
\newcommand{\mean}{\operatorname{mean}}
\newcommand{\trace}{\operatorname{tr}}
\newcommand{\diag}{\operatorname{diag}}
\newcommand{\vect}{\operatorname{vec}}
\newcommand{\cov}{\operatorname{cov}}
\newcommand{\sign}{\operatorname{sign}}
\newcommand{\prj}{\operatorname{proj}}

\newcommand{\softmax}{\operatorname{softmax}}
\newcommand{\clip}{\operatorname{clip}}

\newcommand{\defn}{\mathrel{:=}}
\newcommand{\peq}{\mathrel{+\!=}}
\newcommand{\meq}{\mathrel{-\!=}}

\newcommand{\floor}[1]{\left\lfloor{#1}\right\rfloor}
\newcommand{\ceil}[1]{\left\lceil{#1}\right\rceil}
\newcommand{\inner}[1]{\left\langle{#1}\right\rangle}
\newcommand{\norm}[1]{\left\|{#1}\right\|}
\newcommand{\abs}[1]{\left|{#1}\right|}
\newcommand{\frob}[1]{\norm{#1}_F}
\newcommand{\card}[1]{\left|{#1}\right|\xspace}

\newcommand{\diff}{\mathrm{d}}
\newcommand{\der}[3][]{\frac{\diff^{#1}#2}{\diff#3^{#1}}}
\newcommand{\ider}[3][]{\diff^{#1}#2/\diff#3^{#1}}
\newcommand{\pder}[3][]{\frac{\partial^{#1}{#2}}{\partial{{#3}^{#1}}}}
\newcommand{\ipder}[3][]{\partial^{#1}{#2}/\partial{#3^{#1}}}
\newcommand{\dder}[3]{\frac{\partial^2{#1}}{\partial{#2}\partial{#3}}}

\newcommand{\wb}[1]{\overline{#1}}
\newcommand{\wt}[1]{\widetilde{#1}}

\def\xssp{\hspace{1pt}}
\def\ssp{\hspace{3pt}}
\def\msp{\hspace{5pt}}
\def\lsp{\hspace{12pt}}

\newcommand{\cA}{\mathcal{A}}
\newcommand{\cB}{\mathcal{B}}
\newcommand{\cC}{\mathcal{C}}
\newcommand{\cD}{\mathcal{D}}
\newcommand{\cE}{\mathcal{E}}
\newcommand{\cF}{\mathcal{F}}
\newcommand{\cG}{\mathcal{G}}
\newcommand{\cH}{\mathcal{H}}
\newcommand{\cI}{\mathcal{I}}
\newcommand{\cJ}{\mathcal{J}}
\newcommand{\cK}{\mathcal{K}}
\newcommand{\cL}{\mathcal{L}}
\newcommand{\cM}{\mathcal{M}}
\newcommand{\cN}{\mathcal{N}}
\newcommand{\cO}{\mathcal{O}}
\newcommand{\cP}{\mathcal{P}}
\newcommand{\cQ}{\mathcal{Q}}
\newcommand{\cR}{\mathcal{R}}
\newcommand{\cS}{\mathcal{S}}
\newcommand{\cT}{\mathcal{T}}
\newcommand{\cU}{\mathcal{U}}
\newcommand{\cV}{\mathcal{V}}
\newcommand{\cW}{\mathcal{W}}
\newcommand{\cX}{\mathcal{X}}
\newcommand{\cY}{\mathcal{Y}}
\newcommand{\cZ}{\mathcal{Z}}

\newcommand{\vA}{\mathbf{A}}
\newcommand{\vB}{\mathbf{B}}
\newcommand{\vC}{\mathbf{C}}
\newcommand{\vD}{\mathbf{D}}
\newcommand{\vE}{\mathbf{E}}
\newcommand{\vF}{\mathbf{F}}
\newcommand{\vG}{\mathbf{G}}
\newcommand{\vH}{\mathbf{H}}
\newcommand{\vI}{\mathbf{I}}
\newcommand{\vJ}{\mathbf{J}}
\newcommand{\vK}{\mathbf{K}}
\newcommand{\vL}{\mathbf{L}}
\newcommand{\vM}{\mathbf{M}}
\newcommand{\vN}{\mathbf{N}}
\newcommand{\vO}{\mathbf{O}}
\newcommand{\vP}{\mathbf{P}}
\newcommand{\vQ}{\mathbf{Q}}
\newcommand{\vR}{\mathbf{R}}
\newcommand{\vS}{\mathbf{S}}
\newcommand{\vT}{\mathbf{T}}
\newcommand{\vU}{\mathbf{U}}
\newcommand{\vV}{\mathbf{V}}
\newcommand{\vW}{\mathbf{W}}
\newcommand{\vX}{\mathbf{X}}
\newcommand{\vY}{\mathbf{Y}}
\newcommand{\vZ}{\mathbf{Z}}

\newcommand{\va}{\mathbf{a}}
\newcommand{\vb}{\mathbf{b}}
\newcommand{\vc}{\mathbf{c}}
\newcommand{\vd}{\mathbf{d}}
\newcommand{\ve}{\mathbf{e}}
\newcommand{\vf}{\mathbf{f}}
\newcommand{\vg}{\mathbf{g}}
\newcommand{\vh}{\mathbf{h}}
\newcommand{\vi}{\mathbf{i}}
\newcommand{\vj}{\mathbf{j}}
\newcommand{\vk}{\mathbf{k}}
\newcommand{\vl}{\mathbf{l}}
\newcommand{\vm}{\mathbf{m}}
\newcommand{\vn}{\mathbf{n}}
\newcommand{\vo}{\mathbf{o}}
\newcommand{\vp}{\mathbf{p}}
\newcommand{\vq}{\mathbf{q}}
\newcommand{\vr}{\mathbf{r}}
\newcommand{\Vs}{\mathbf{s}}
\newcommand{\vt}{\mathbf{t}}
\newcommand{\vu}{\mathbf{u}}
\newcommand{\vv}{\mathbf{v}}
\newcommand{\vw}{\mathbf{w}}
\newcommand{\vx}{\mathbf{x}}
\newcommand{\vy}{\mathbf{y}}
\newcommand{\vz}{\mathbf{z}}

\newcommand{\vone}{\mathbf{1}}
\newcommand{\vzero}{\mathbf{0}}

\newcommand{\valpha}{{\boldsymbol{\alpha}}}
\newcommand{\vbeta}{{\boldsymbol{\beta}}}
\newcommand{\vgamma}{{\boldsymbol{\gamma}}}
\newcommand{\vdelta}{{\boldsymbol{\delta}}}
\newcommand{\vepsilon}{{\boldsymbol{\epsilon}}}
\newcommand{\vzeta}{{\boldsymbol{\zeta}}}
\newcommand{\veta}{{\boldsymbol{\eta}}}
\newcommand{\vtheta}{{\boldsymbol{\theta}}}
\newcommand{\viota}{{\boldsymbol{\iota}}}
\newcommand{\vkappa}{{\boldsymbol{\kappa}}}
\newcommand{\vlambda}{{\boldsymbol{\lambda}}}
\newcommand{\vmu}{{\boldsymbol{\mu}}}
\newcommand{\vnu}{{\boldsymbol{\nu}}}
\newcommand{\vxi}{{\boldsymbol{\xi}}}
\newcommand{\vomikron}{{\boldsymbol{\omikron}}}
\newcommand{\vpi}{{\boldsymbol{\pi}}}
\newcommand{\vrho}{{\boldsymbol{\rho}}}
\newcommand{\vsigma}{{\boldsymbol{\sigma}}}
\newcommand{\vtau}{{\boldsymbol{\tau}}}
\newcommand{\vupsilon}{{\boldsymbol{\upsilon}}}
\newcommand{\vphi}{{\boldsymbol{\phi}}}
\newcommand{\vchi}{{\boldsymbol{\chi}}}
\newcommand{\vpsi}{{\boldsymbol{\psi}}}
\newcommand{\vomega}{{\boldsymbol{\omega}}}

\newcommand{\rLambda}{\mathrm{\Lambda}}
\newcommand{\rSigma}{\mathrm{\Sigma}}

\newcommand{\vLambda}{\bm{\rLambda}}
\newcommand{\vSigma}{\bm{\rSigma}}

\makeatletter
\newcommand*\bdot{\mathpalette\bdot@{.7}}
\newcommand*\bdot@[2]{\mathbin{\vcenter{\hbox{\scalebox{#2}{$\m@th#1\bullet$}}}}}
\makeatother

\makeatletter
\DeclareRobustCommand\onedot{\futurelet\@let@token\@onedot}
\def\@onedot{\ifx\@let@token.\else.\null\fi\xspace}

\def\eg{\emph{e.g}\onedot} \def\Eg{\emph{E.g}\onedot}
\def\ie{\emph{i.e}\onedot} \def\Ie{\emph{I.e}\onedot}
\def\cf{\emph{cf}\onedot} \def\Cf{\emph{Cf}\onedot}
\def\etc{\emph{etc}\onedot} \def\vs{\emph{vs}\onedot}
\def\wrt{w.r.t\onedot} \def\dof{d.o.f\onedot} \def\aka{a.k.a\onedot}
\def\etal{\emph{et al}\onedot}
\makeatother

\newcommand{\relu}{\operatorname{ReLU}}
\newcommand{\gap}{\operatorname{GAP}}
\newcommand{\up}{\operatorname{Up}}
\newcommand{\ce}{\operatorname{CE}}

\newcommand{\cam}{\textrm{CAM}}
\newcommand{\gcam}{\textrm{Grad-CAM}}
\newcommand{\scam}{\textrm{Score-CAM}}

\newcommand{\mae}{\textrm{MAE}}
\newcommand{\mse}{\textrm{MSE}}
\newcommand{\hi}{\textrm{HI}}

\title{A Learning Paradigm for Interpretable Gradients}


\author{\authorname{Felipe Torres Figueroa\sup{1}, Hanwei Zhang\sup{1}, Ronan Sicre\sup{1}, Yannis Avrithis\sup{2} and Stephane Ayache\sup{1}}
\affiliation{\sup{1}Centrale Marseille, Aix Marseille Univ, CNRS, LIS, Marseille, France}
\affiliation{\sup{2}Institute of Advanced Research on Artificial Intelligence (IARAI)}
\email{\{felipe.torres, hanwei.zhang, ronan.sicre, stephane.ayache\}@lis-lab.fr}
}


\keywords{Gradient, Guided Backpropagation, Class Activation Maps, Interpretability}

\abstract{This paper
studies interpretability of convolutional networks by means of saliency maps. Most approaches based on Class Activation Maps (CAM) combine information from fully connected layers and gradient through variants of backpropagation. However, it is well understood that gradients are noisy and alternatives like guided backpropagation have been proposed to obtain better visualization at inference. In this work, we present a novel training approach to improve the quality of gradients for interpretability. In particular, we introduce a regularization loss such that the gradient with respect to the input image obtained by standard backpropagation is similar to the gradient obtained by guided backpropagation. We find that the resulting gradient is qualitatively less noisy and improves quantitatively the interpretability properties of different networks, using several interpretability methods.}

\onecolumn \maketitle \normalsize \setcounter{footnote}{0} \vfill

\uppercase{\section{Introduction}}
\label{sec:intro}

The improvement of deep learning models in the last decade has led to their adoption and penetration into most application sectors. Since these models are highly complex and opaque, the requirement for interpretability of their predictions receives a lot of attention~\cite{mythos_interp}. Explanation and transparency becomes a legal requirements for systems used in high-stakes and high-risk decisions.

In this work, we focus on the visual interpretability of deep learning models. Model interpretability is often categorized into \emph{transparency} and \emph{post-hoc} methods. Transparency aims at producing models where the inner process or part of it can be understood. Post-hoc methods consider models as black-boxes and interpret decisions mainly based on inputs and outputs.

In visual recognition,
most methods focus on post-hoc interpretability by means of \emph{saliency maps}. These maps highlight the most important areas of an image related to the network prediction. Initial works focused on using gradients for visualization, such as guided backpropagation~\cite{guidedbackprop}. CAM~\cite{cam} later proposed a class-specific linear combination of feature maps and opened the way to numerous weighting strategies.

Most CAM-based methods use backpropagation in one way or another. Recognizing that the gradients obtained this way are noisy, methods like SmoothGrad~\cite{smoothgrad} and SmoothGrad-CAM++~\cite{omeiza2019smooth} improve the quality of saliency maps by denoising the gradients. However, this requires several forward passes, thus comes with increased cost at inference.

In this work, we rather propose a \emph{learning paradigm} for model training that regularizes gradients to improve the performance of interpretability methods. In particular, we add a regularization term to the loss function that encourages the gradient in the input space to align with the gradient obtained by guided back-propagation.
This has a smoothing effect on gradient and is shown to improve the power of model interpretations.

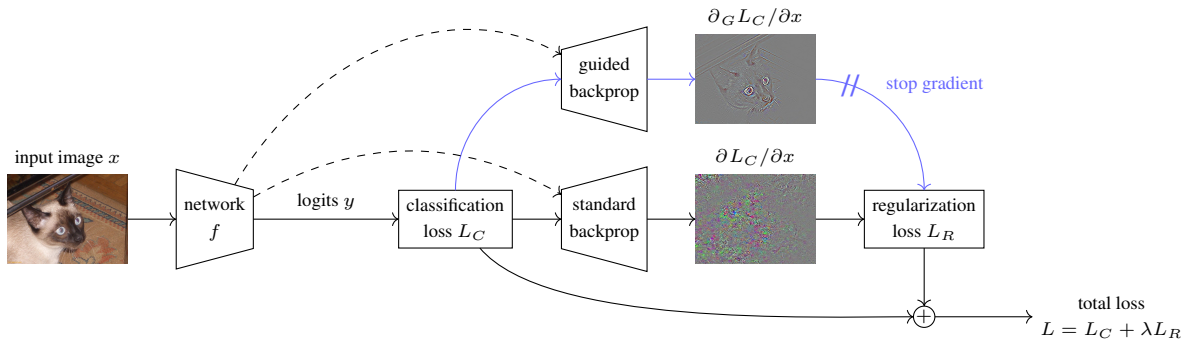
\begin{figure*}[t]
\centering

\begin{tikzpicture}[
	scale=.3,
	font={\scriptsize},
	node distance=.5,
	label distance=3pt,
	wide/.style={yscale=.75},
	net/.style={draw,trapezium,trapezium angle=75,inner sep=3pt},
	enc/.style={net,shape border rotate=270},
	dec/.style={net,shape border rotate=90},
	txt/.style={inner sep=3pt},
	loose/.style={looseness=.6},
	sg/.style={blue!60},
]
\matrix[
	tight,row sep=6,column sep=18,
	cells={scale=.3,},
] {
	\&\&\&\&\&
	\node[dec] (guided-back) {guided \\ backprop}; \&
	\node[wide,label=90:$\ipder{_G L_C}{x}$] (guided) {\fig[.1]{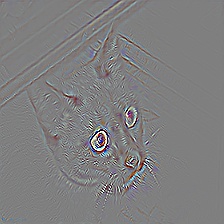}}; \\
	\node[label=90:input image $x$] (in) {\fig[.1]{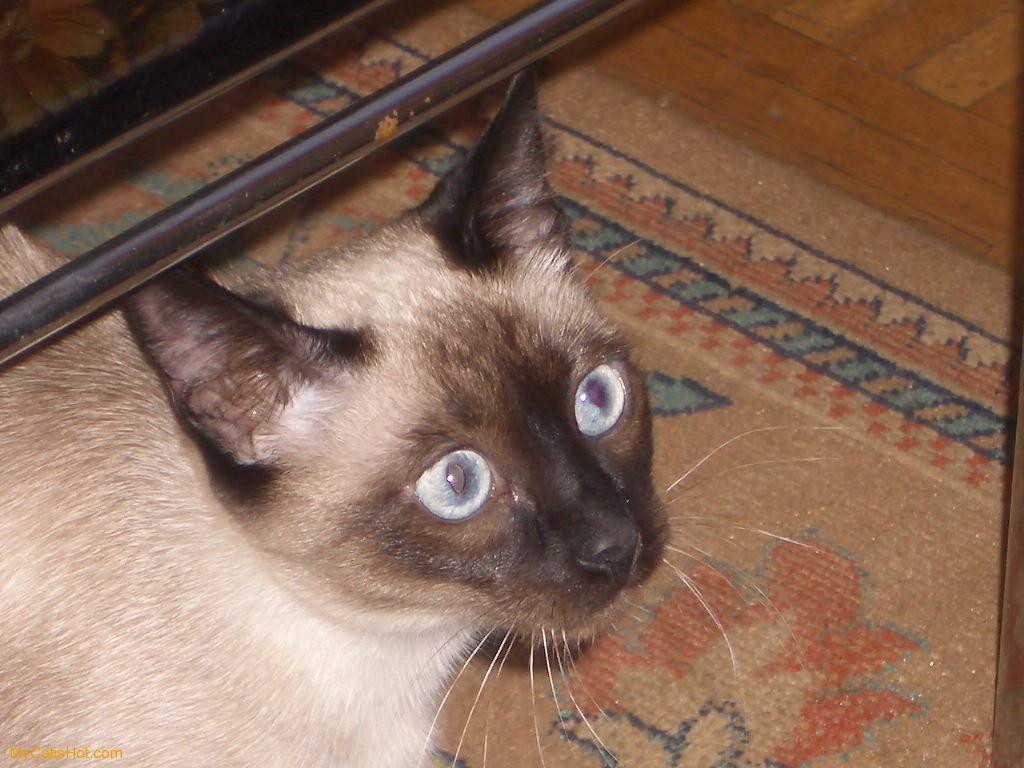}}; \&
	\node[enc] (net) {network \\ $f$}; \&\&\&
	\node[box] (class) {classification \\ loss $L_C$}; \&
	\node[dec] (back) {standard \\ backprop}; \&
	\node[wide,label=90:$\ipder{L_C}{x}$] (grad) {\fig[.1]{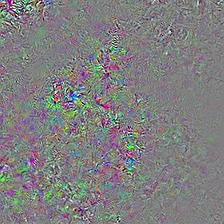}}; \&
	\node[box] (reg) {regularization \\ loss $L_R$}; \\
	\&\&\&\&\&\&\&
	\node[op] (plus) {$+$}; \&
	\node[txt] (loss) {total loss \\ $L = L_C + \lambda L_R$}; \\
};

\draw[->]
	(in) edge (net)
	(net) edge node[above] {logits $y$} (class)
	(class) edge (back)
	(back) edge (grad)
	(guided-back) edge[sg] (guided)
	(grad) edge (reg)
	(reg) edge (plus)
	(plus) edge (loss)
	(class) edge[sg,out=90,in=180] (guided-back)
	(class) edge[loose,out=-50,in=180] (plus)
	(guided) edge[sg,out=0,in=90] node[pos=.2,font=\large,label=0:stop gradient] {//} (reg)
	(net) edge[dashed,out=60,in=150] (guided-back)
	(net) edge[dashed,out=30,in=150] (back)
	;

\end{tikzpicture}

\caption{\emph{Interpretable gradient learning}. For an input image $x$, we obtain the logit vector $y = f(x; \theta)$ by a forward pass through the network $f$ with parameters $\theta$. We compute the classification loss $L_C$ by softmax and cross-entropy~\eq{class}, \eq{ce}. We obtain the standard gradient $\ipder{L_C}{x}$ and guided gradient $\ipder{_G L_C}{x}$ by two backward passes (dashed) and compute the regularization loss $L_R$ as the error between the two~\eq{reg},\eq{mae}-\eq{cos}. The total loss is $L = L_C + \lambda L_R$~\eq{total}. Learning is based on $\ipder{L}{\theta}$, which involves differentiation of the entire computational graph except the guided backpropagation branch (blue).}
\label{fig:method}
\end{figure*}

\autoref{fig:method} summarizes our method. At training, each input image is forwarded through the network to compute the cross-entropy loss. Standard and guided backpropagation is performed back to the input image space, where our regularization term is computed. This term is added to the loss and backpropagated only through the standard backpropagation branch.

The key contributions of this work are as follows:
\begin{itemize}
    \item We introduce a new learning paradigm to regularize gradients.
    \item Using different networks, we show that our method improves the gradient quality and the performance of several interpretability methods using multiple metrics, while 
    preserving accuracy.
\end{itemize}

\section{\uppercase{Related Work}}
\label{sec:related}

Interpretability of deep neural network decisions is a problem that receives increasing interest.
As interpretability is not simple to define, Lipton~\cite{mythos_interp} proposes some common ground, definitions and categorization for interpretability methods. For instance, \emph{transparency} aims at making models simple so it is humanly possible to provide an explanation of its inner mechanism. By contrast, \emph{post-hoc} methods consider models as black boxes and study the activations leading to a specific output.

LIME~\cite{LIME} and SHAP~\cite{NIPS2017_7062} are probably the most popular post-hoc methods that are model agnostic and provide local information. Concerning image recognition tasks, it is common to generate \emph{saliency maps} highlighting the areas of an image that are responsible for a specific prediction. Several of these methods are either based on backpropagation and its variants or on Class Activation Maps (CAM) that weigh the importance of activation maps.


\subsection{Gradient-based approaches}

Gradient-based approaches assess the impact of distinct image regions on the prediction based on the partial derivative of the model prediction function with respect to the input. A simple saliency map can be the partial derivative obtained by a single backward pass through the model~\cite{simonyan2013deep}.

\emph{Guided backpropagation}~\cite{guidedbackprop} enhances explanations by removing negative gradients through ReLU units. For better visualization, \emph{SmoothGrad}~\cite{smoothgrad} applies noise to the input and derives saliency maps based on the average of resulting gradients. \emph{Layer-wise Relevance Propagation (LRP)}~\cite{bach2015pixel} reallocates the prediction score through a custom backward pass across the network.

Our method has a similar objective as \emph{SmoothGrad}~\cite{smoothgrad} but instead of using several forward passes at inference, we regularize gradients using guided backpropagation during training. Thus we obtain better gradients without modifying the inference process and our method can be used with any interpretability method at inference.


\subsection{CAM-based approaches}

Class Activation Maps~\cite{cam} produces a saliency map that highlights the areas of an image that are the most responsible for a CNN decision. The saliency map is computed as a linear combinations of feature maps from a given layer. Different variants of CAM are proposed by defining different weighting coefficients. Grad-CAM~\cite{gradcam}, for instance, spatially averages the gradient with respect to feature maps. Grad-CAM++~\cite{gradcampp} improves object localization by using positive partial derivatives and measuring recognition and localization metrics.

It is possible to extend CAM to multiple layers~\cite{layercam} and to improve sensitivity~\cite{sundararajan2017axiomatic} and conservation~\cite{montavon2018methods} by the addition of axioms~\cite{axiombased}. Score-CAM~\cite{scorecam} is a gradient-free method that computes weighting coefficients by maximizing the Average Increase metric~\cite{gradcampp}. Further improvement can be obtained by means of test-time optimization~\cite{zhang2023opti}.

Some works provide explanations that not only localize salient parts of images, but also provide theoretical bases on the effect of modifying such regions for a given input~\cite{axiombased}. An exhaustive alternative performs ablation experiments to highlight such parts~\cite{ramaswamy2020ablation}.

All these approaches apply at inference, without modifying the model or the training process. By contrast, our work applies at training with the objective of improving the quality of gradients, which is much needed for gradient-based methods. Thus, our method is orthogonal and can be used with any of these approaches at inference.


\subsection{Double backpropagation}

Double backpropagation is a general regularization paradigm, first introduced by Drucker and Le Cun~\cite{drucker1991double} to improve generalization. The idea is used to avoid overfitting~\cite{philipp2018nonlinearity}, help transfer~\cite{srinivas2018knowledge}, cope with noisy labels~\cite{luo2019simple}, and more recently to increase adversarial robustness~\cite{lyu2015unified,simon2018adversarial,ross2018improving,seck20191,finlay2018improved}.
It aims at penalizing the $\ell_1$~\cite{seck20191}, $\ell_2$ or $\ell_\infty$ norm of the gradient with respect to the input image.

Our method is related and regularizes the standard gradient by aligning it with the guided gradient, obtained by guided backpropagation~\cite{guidedbackprop}.

\uppercase{\section{Background}}
\label{sec:back}


\subsection{Guided backpropagation}

The derivative of $v = \relu(u) = [u]_+ = \max(u,0)$ with respect to $u$ is $\ider{v}{u} = \ind_{u>0}$. By the chain rule, a signal $\delta v = \ipder{L}{v}$ is then propagated backwards through the $\relu$ unit to $\delta u = \ipder{L}{u}$ as $\delta u = \ind_{u>0} \delta v$, where $\ipder{L}{v}$ is the partial derivative of any scalar quantity of interest, \eg a loss $L$, with respect to $v$.

\emph{Guided backpropagation}~\cite{guidedbackprop} changes this to $\delta_G u = \ind_{u>0} [\delta v]_+$, masking out values corresponding to negative entries of both the forward ($u$) and the backward ($\delta v$) signals and thus preventing backward flow of negative gradients.

Standard backpropagation through an entire network $f$ with this particular change for $\relu$ units is called \emph{guided backpropagation}. The corresponding guided ``partial derivative'' or \emph{guided gradient} of scalar quantity $L$ with respect to $v$ is denoted by $\ipder{_G L}{v}$. This method allows sharp visualization of high-level activations conditioned on input images.


\subsection{CAM-based methods}

CAM-based methods build a saliency map as a linear combination of feature maps. Given a target class $c$ and a set of 2D feature maps $\{A^k\}_{k=1}^K$, the \emph{saliency map} is defined as
\begin{equation}
	S^c = \relu \left( \sum_{k=1}^K \alpha^c_k A^k \right),
\label{eq:cam}
\end{equation}
where the weight $\alpha^c_k$ determines the contribution of channel $k$ to class $c$. The saliency map $S^c$ and the feature maps $A^k$ are both non-negative because of using $\relu$ activation functions. Different CAM-based methods differ primarily in the definition of the weights $\alpha^c_k$.


\paragraph{CAM~\cite{cam}}

originally defines $\alpha^c_k$ as the weight connecting channel $k$ to class $c$ in the classifier, assuming $\{A^k\}$ are the feature maps of the last convolutional layer, which is followed by \emph{global average pooling} (GAP) and a fully connected layer.


\paragraph{Grad-CAM~\cite{gradcam}}

is a generalization of CAM for any network. If $y^c$ is the logit of class $c$, the weights are obtained by GAP of the partial derivatives of $y^c$ with respect to elements of feature map $A^k$ of any given layer,
\begin{equation}
	\alpha^c_k = \frac{1}{Z} \sum_{i,j} \pder{y^c}{A^k_{ij}},
\label{eq:gcam}
\end{equation}
where $A^k_{ij}$ denotes the value at spatial location $(i,j)$ of feature map $A^k$ and $Z$ is the total number of locations.

Guided Grad-CAM elementwise-multiplies the saliency maps obtained by Grad-CAM and guided backpropagation, after adjusting spatial resolutions. The resulting visualizations are both class-discriminative (by Grad-CAM) and contain fine-grained detail (by guided backpropagation).


\paragraph{Grad-CAM++~\cite{gradcampp}}

is a generalization of Grad-CAM, where partial derivatives of $y^c$ with respect to $A^k$ are followed by $\relu$ as in guided backpropagation and GAP is replaced by a weighted average:
\begin{equation}
    a^c_k = \sum_{i,j} w^{kc}_{ij} \relu \left( \pder{y^c}{A^k_{ij}} \right).
\label{eq:gcamp}
\end{equation}
The weights $w_{ij}^{kc}$ of the linear combination are
\begin{equation}
    w_{ij}^{kc} = \frac{\pder[2]{y^c}{(A^k_{ij})}}
		{2\pder[2]{y^c}{(A^k_{ij})} + \sum_{a,b} A^k_{ab}\pder[3]{y^c}{(A^k_{ij})}}.
\label{eq:gcampw}
\end{equation}


%


\paragraph{Score-CAM~\cite{scorecam}}

computes the weights $a^c_k$ based on the increase in confidence~\cite{gradcampp} for class $c$ obtained by masking (element-wise multiplying) the input image $x$ with feature map $A^k$:
\begin{equation}
	a^c_k = f(x \circ s(\up(A^k)))^c - f(x_b)^c,
\label{eq:scam}
\end{equation}
where $\up$ is upsampling to the spatial resolution of $x$, $s$ is linear normalization to range $[0,1]$, $\circ$ is the Hadamard product, $f$ is the network mapping of input image to class probability vectors and $x_b$ is a baseline image.

While Score-CAM does not require gradients to compute saliency maps, \eq{scam} requires one forward pass through the network $f$ for each channel $k$.

\uppercase{\section{Method}}
\label{sec:method}

\paragraph{Preliminaries}

We consider an image classification network $f$ with parameters $\theta$, which maps an input image $x$ to a vector of predicted class probabilities $p = f(x; \theta)$. At inference, we predict the class of maximum confidence $\arg\max_j p^j$, where $p^j$ is the probability of class $j$. At training, given training images $X = \{x_i\}_{i=1}^n$ and target labels $T = \{t_i\}_{i=1}^n$, we compute the \emph{classification loss}
\begin{equation}
	L_C(X, \theta, T) = \frac{1}{n} \sum_{i=1}^n \ce(f(x_i; \theta), t_i),
\label{eq:class}
\end{equation}
where $\ce$ is cross-entropy:
\begin{equation}
	\ce(p, t) = -\log p^t.
\label{eq:ce}
\end{equation}
Updates of parameters $\theta$ are performed by an optimizer, based on the standard partial derivative (gradient) $\ipder{L_C}{\theta}$ of the classification loss $L_C$ with respect to $\theta$,  obtained by standard back-propagation.


\paragraph{Motivation}

Due to non-linearities like ReLU activations and downsampling like max-pooling or convolution stride greater than $1$, the standard gradient is noisy~\cite{smoothgrad}. This is shown by visualizing the gradient $\ipder{L_C}{x}$ with respect to an input image $x$. By contrast, the guided gradient $\ipder{_G L_C}{x}$ \cite{guidedbackprop} does not suffer much from noise and preserves sharp details. The difference of the two gradients is illustrated in \autoref{fig:method}.

The main motivation of this work is that introducing a regularization term during training could make the standard gradient $\ipder{L_C}{x}$ behave similarly to the corresponding guided gradient $\ipder{_G L_C}{x}$, while maintaining the predictive power of the classifier $f$. We hypothesize that, if this is possible, it will improve the quality of all gradients with respect to intermediate activations and therefore the quality of saliency maps obtained by CAM-based methods \cite{cam,gradcam,gradcampp,scorecam} and the interpretability of network $f$. The effect may be similar to that of SmoothGrad \cite{smoothgrad}, but without the need for several forward passes at inference.


\paragraph{Regularization}

Given an input training image $x_i$ and its target labels $t_i$, we perform a forward pass through $f$ and compute the probability vectors $p_i = f(x_i, \theta)$ and the contribution of $(x_i, t_i)$ to the classification loss $L_C(X, \theta, T)$~\eq{class}.

We then obtain the standard gradients $\delta x_i = \ipder{L_C}{x_i}$ and the guided gradients $\delta_G x_i = \ipder{_G L_C}{x_i}$ with respect to $x_i$ by two separate backward passes. Since the whole process is differentiable (\wrt $\theta$) at training, we stop the gradient computation of the latter, so that it only serves as a ``teacher''. We define the \emph{regularization loss}
\begin{equation}
	L_R(X, \theta, T) = \frac{1}{n} \sum_{i=1}^n E(\delta x_i, \delta_G x_i),
\label{eq:reg}
\end{equation}
where $E$ is an error function between the two gradient images, considered below.

The total loss is defined as
\begin{equation}
	L(X, \theta, T) = L_C(X, \theta, T) + \lambda L_R(X, \theta, T),
\label{eq:total}
\end{equation}
where $\lambda$ is a hyperparameter determining the regularization strength. $\lambda$ should be large enough to smooth the gradient without decreasing the classification accuracy or harming the training process.

Updates of the network parameters $\theta$ are based on the standard gradient $\ipder{L}{\theta}$ of the total loss $L$ \wrt $\theta$, using any optimizer. At inference, one may use any interpretability method to obtain a saliency map at any layer.


\paragraph{Error function}

Given two gradient images $\delta, \delta'$ consisting of $m$ pixels each, we consider the following error functions $E$ to compute the regularization loss~\eq{reg}.
\begin{enumerate}[itemsep=2pt, parsep=0pt, topsep=3pt]
	\item \emph{Mean absolute error} (MAE):
	\begin{equation}
		E_\mae(\delta, \delta') = \frac{1}{m} \norm{\delta - \delta'}_1.
	\label{eq:mae}
	\end{equation}
	\item \emph{Mean squared error} (MSE):
	\begin{equation}
		E_\mse(\delta, \delta') = \frac{1}{m} \norm{\delta - \delta'}_2^2.
	\label{eq:mse}
	\end{equation}
\end{enumerate}

We also consider the following two similarity functions, with a negative sign.
\begin{enumerate}[itemsep=2pt, parsep=0pt, topsep=3pt]
	\setcounter{enumi}{2}
	\item \emph{Cosine similarity}:
	\begin{equation}
		E_{\cos}(\delta, \delta') = -\frac{\inner{\delta, \delta'}}{\norm{\delta}_2 \norm{\delta'}_2},
	\label{eq:cos}
	\end{equation}
	where $\inner{,}$ denotes inner product.
	\item \emph{Histogram intersection} (HI):
	\begin{equation}
		E_\hi(\delta, \delta') = -\frac
			{\sum_{i=0}^m \min(\abs{\delta_i}, \abs{\delta'_i})}
			{\norm{\delta}_1 \norm{\delta'}_1}.
	\label{eq:hi}
	\end{equation}
\end{enumerate}


\paragraph{Algorithm}

Our method is summarized in \autoref{alg:grad} and illustrated in \autoref{fig:method}. It is interesting to note that the entire computational graph depicted in \autoref{fig:method} involves one forward and two backward passes. This graph is then differentiated again to compute $\ipder{L}{\theta}$, which involves one more forward and backward pass, since the guided backpropagation branch is excluded. Thus, each training iteration requires five passes through $f$ instead of two in standard training.

\begin{algorithm}
	\SetFuncSty{textsc}
	\SetDataSty{emph}
	\newcommand{\commentsty}[1]{{\color{DarkGreen}#1}}
	\SetCommentSty{commentsty}
	\SetKwComment{Comment}{$\triangleright$ }{}

	\SetKwFunction{Detatch}{detatch}

	\KwIn{network $f$, parameters $\theta$}
	\KwIn{input images $X = \{x_i\}_{i=1}^n$}
	\KwIn{target labels $T = \{t_i\}_{i=1}^n$}
	\KwOut{loss $L$}
	$L_C \gets \frac{1}{n} \sum_{i=1}^n \ce(f(x_i; \theta), t_i)$ \Comment*[f]{class. loss \eq{class}} \\
	\ForEach{$i \in \{1,\dots,n\}$}{
		$\delta x_i \gets \ipder{L_C}{x_i}$ \Comment*[f]{standard grad} \\
		$\delta_G x_i \gets \ipder{_G L_C}{x_i}$ \Comment*[f]{guided grad} \\
		$\mbox{\Th{Detach}}(\delta_G x_i)$ \Comment*[f]{detach from graph}
	}
	$L_R \gets \frac{1}{n} \sum_{i=1}^n E(\delta x_i, \delta_G x_i) $ \Comment*[f]{reg. loss \eq{reg}} \\
	$L \gets L_C + \lambda L_R$ \Comment*[f]{total loss ~\eq{total}} \\

\caption{Interpretable gradient loss}
\label{alg:grad}
\end{algorithm}

\uppercase{\section{Experiments}}
\label{sec:exp}

\subsection{Experimental setup}

In the following sections, we evaluate the effect of our approach on recognition and interpretability.

\paragraph{Models and datasets}

We train and evaluate a ResNet-18~\cite{he16} and a MobileNet-V2~\cite{sandler2018mobilenetv2} on CIFAR-100~\cite{krizhevsky2009learning}. ResNets are the most common CNNs and the ResNet-18 is particularly adapted to low resolution images. MobileNet-V2 is a widely used compact CNN. CIFAR-100 contains 60.000 images of 100 categories, split in 50.000 for training and 10.000 for testing. Each image has a resolution of $32\times32$ pixels.

\paragraph{Settings}

To obtain competitive performance and ensure the replicability of our method, we follow the methodology by weiaicunzai\footnote{https://github.com/weiaicunzai/pytorch-cifar100}. In particular, we train for 200 epochs, with a batch-size of 128 images, SGD optimizer, initial learning rate $10^{-1}$ and learning rate decay by a factor of 5 on epochs 60, 120 and 160.

At inference, we generate explanations following popular attribution methods derived from CAM~\cite{cam}, from the \emph{pytorch-grad-cam} library from Jacob Gildenblat\footnote{https://github.com/jacobgil/pytorch-grad-cam}.

\begin{figure*}
\scriptsize
\centering
\setlength{\tabcolsep}{1.5pt}
\begin{tabular}{cccccccccc}
	& \mr{2}{\Th{Input}} & \mc{2}{\Th{Grad-CAM}} & \mc{2}{\Th{Grad-CAM++}} & \mc{2}{\Th{Score-CAM}} & \mc{2}{\Th{Ablation-CAM}} \\
	& & \Th{Baseline} & \Th{Ours} & \Th{Baseline} & \Th{Ours} & \Th{Baseline} & \Th{Ours} & \Th{Baseline} & \Th{Ours} \\

 	{\rotatebox{90}{\scriptsize Man}} &
 	\fig[.09]{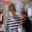} &
 	\fig[.09]{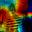} &
 	\fig[.09]{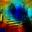} &
 	\fig[.09]{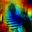} &
 	\fig[.09]{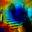} &
 	\fig[.09]{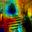} &
 	\fig[.09]{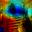} &
 	\fig[.09]{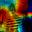} &
 	\fig[.09]{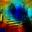} \\

	{\rotatebox{90}{\scriptsize Girl}} &
	\fig[.09]{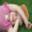} &
	\fig[.09]{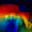} &
	\fig[.09]{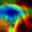} &
	\fig[.09]{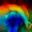} &
	\fig[.09]{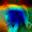} &
	\fig[.09]{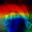} &
	\fig[.09]{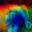} &
	\fig[.09]{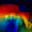} &
	\fig[.09]{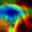} \\

	{\rotatebox{90}{\scriptsize Woman}} &
	\fig[.09]{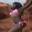} &
	\fig[.09]{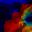} &
	\fig[.09]{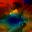} &
	\fig[.09]{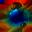} &
	\fig[.09]{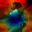} &
	\fig[.09]{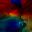} &
	\fig[.09]{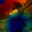} &
	\fig[.09]{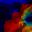} &
	\fig[.09]{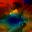}\\

	{\rotatebox{90}{\scriptsize Lobster}} &
	\fig[.09]{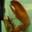} &
	\fig[.09]{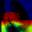} &
	\fig[.09]{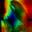} &
	\fig[.09]{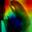} &
	\fig[.09]{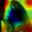} &
	\fig[.09]{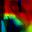} &
	\fig[.09]{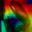} &
	\fig[.09]{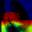} &
	\fig[.09]{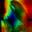} \\

	{\rotatebox{90}{\scriptsize Couch}} &
	\fig[.09]{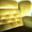} &
	\fig[.09]{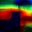} &
	\fig[.09]{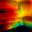} &
	\fig[.09]{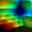} &
	\fig[.09]{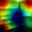} &
	\fig[.09]{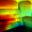} &
	\fig[.09]{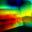} &
	\fig[.09]{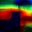} &
	\fig[.09]{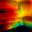} \\
\end{tabular}
\caption{\emph{Saliency map comparison} of standard \vs our training using different CAM-based methods on CIFAR-100 examples.}
\label{fig:salient}
\end{figure*}

\subsection{Faithfulness metrics}

Faithfulness evaluation~\cite{gradcampp} offers insight on the regions of an image that are considered important for recognition, as highlighted by the saliency map $S^c$. Specifically, given a target class $c$, an image $x$ and a saliency map $S^c$ are element-wise multiplied to obtain a \emph{masked image}
\begin{equation*}
    m^c = S^c\circ x.
\end{equation*}
This masked image is similar to the original image on the salient areas and black on the non-salient ones. To evaluate the quality of saliency maps, we forward both the original image $x$ and its masked version $m^c$ through the network to obtain the predicted probabilities $p_i^c$ and $o_i^c$ respectively. We then compute a number of metrics as defined below.

\paragraph{Average Drop (AD)}

aims at quantifying how much predictive power is lost when we consider the masked image compared to the original one. Lower is better.
\begin{equation}
	\mathrm{AD} = \frac{1}{N} \sum_{i=1}^N \frac{[p_i^c- o_i^c]_+}{p_i^c}.
\label{eq:ad}
\end{equation}

\paragraph{Average Increase (AI)}

is also known as Increase of Confidence and measures the percentage of examples of the dataset where the masked image offers a higher probability than the original for the target class. Higher is better.
\begin{equation}
	\mathrm{AI} = \frac{1}{N} \sum_i^N \ind{(p_i^c < o_i^c)}.
\label{eq:ai}
\end{equation}

\paragraph{Average Gain (AG)}

is recently introduced in~\cite{zhang2023opti} and designed to be a symmetric complement of AD, replacing AI. It aims at quantifying how much predictive power is gained when we consider the masked image compared to the original one. Higher is better.
\begin{equation}
	\mathrm{AG} = \frac{1}{N} \sum_{i=1}^N \frac{[o_i^c - p_i^c]_+}{p_i^c}.
\label{eq:ag}
\end{equation}


\subsection{Causal metrics}

Causality evaluation~\cite{petsiuk2018rise} aims at evaluating the effect of masking certain elements of the image in the predictive power of a model. Two metrics are defined as follows. Histograms and average values can be computed per image. Following most previous work, we only show average values over the test set.

\paragraph{Insertion}

starts from a blurry image and gradually inserts (unblurs) pixels of the original image, ranked by decreasing saliency as defined in a given saliency map. At each iteration, images are passed through the network to compute the predicted probabilities and compare to the original.

\paragraph{Deletion}

gradually removes the pixels by replacing them by black, starting from the most salient pixels. As for insertion, we compute the predicted probabilities at each iteration.

\begin{figure}
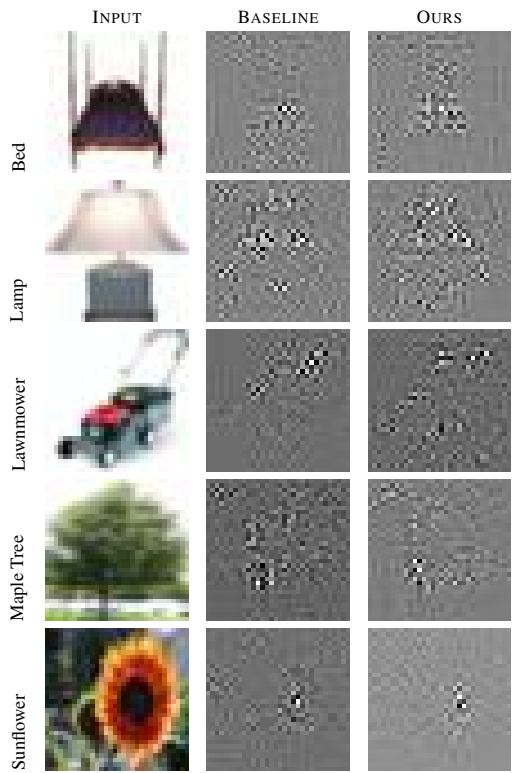

\scriptsize
\centering
\setlength{\tabcolsep}{3.5pt}
\begin{tabular}{cccc}
	& \Th{Input} & \Th{Baseline} & \Th{Ours} \\

	{\rotatebox{90}{\scriptsize Bed}} &
	\fig[.25]{/fig_grad/original/5415} &
	\fig[.25]{/fig_grad/baseline/5415} &
	\fig[.25]{/fig_grad/cosine/5415} \\

 	{\rotatebox{90}{\scriptsize Lamp}} &
 	\fig[.25]{/fig_grad/original/1766} &
 	\fig[.25]{/fig_grad/baseline/1766} &
 	\fig[.25]{/fig_grad/cosine/1766} \\

	{\rotatebox{90}{\scriptsize Lawnmower}} &
	\fig[.25]{/fig_grad/original/8311} &
	\fig[.25]{/fig_grad/baseline/8311} &
	\fig[.25]{/fig_grad/cosine/8311} \\

	{\rotatebox{90}{\scriptsize Maple Tree}} &
	\fig[.25]{/fig_grad/original/1198} &
	\fig[.25]{/fig_grad/baseline/1198} &
	\fig[.25]{/fig_grad/cosine/1198} \\

	{\rotatebox{90}{\scriptsize Sunflower}} &
	\fig[.25]{/fig_grad/original/8881} &
	\fig[.25]{/fig_grad/baseline/8881} &
	\fig[.25]{/fig_grad/cosine/8881} \\
\end{tabular}
\caption{\emph{Gradient comparison} of standard \vs our training on CIFAR-100 examples.}
\label{fig:grads}
\end{figure}

\subsection{Qualitative results}

We visualize the effect of our approach on saliency maps and gradients, obtained for the baseline model \vs the one trained with our approach.

\autoref{fig:salient} shows saliency maps. We observe the differences brought by our training method. The differences are particularly important for Grad-CAM, which directly averages the gradient to weigh feature maps. Interestingly, the differences are smaller for Score-CAM, which is not gradient-based but only obtains changes of predicted probabilities.

\autoref{fig:grads} shows gradients. We observe slightly less noise with our method, while the object of interest is better covered by gradient activations.

\begin{table}
\centering
\scriptsize
\begin{tabular}{lccc} \toprule
	\Th{Model} & \Th{Error} & \Th{$\lambda$}&\Th{Acc} \\ \midrule
	\mr{2}{\Th{ResNet-18}}& Baseline & -- &\Th{73.42} \\
	& Ours & $7.5\times10^{-3}$ & \Th{72.86} \\ \midrule
	\mr{2}{\Th{MobileNet-V2}} & Baseline & -- &\Th{59.43} \\
	& Ours & $1\times10^{-3}$ & \Th{62.36} \\ \midrule
\end{tabular}
\caption{\emph{Accuracy} of standard \vs our training using ResNet-18 and MobileNet-V2 on CIFAR-100. Using cosine error function for our training.}
\label{tab:C100_acc}
\end{table}

\begin{table}
\centering
\scriptsize
\setlength{\tabcolsep}{3pt}
\begin{tabular}{lcccccc} \toprule
	\mc{7}{\Th{ResNet-18}}\\\midrule
	\Th{Method}&\Th{Error}&\Th{AD$\downarrow$}&\Th{AG$\uparrow$}&\Th{AI$\uparrow$}&\Th{Ins$\uparrow$}&\Th{Del$\downarrow$}\\\hline
	\mr{2}{\Th{Grad-CAM}}& Baseline &30.16&15.23&29.99&58.47&17.47\\
	& Ours &28.09&16.19&31.53&58.76&17.57\\\hline
	\mr{2}{\Th{Grad-CAM++}}& Baseline &31.40&14.17&28.47&58.61&17.05\\
		& Ours &29.78&15.07&29.60&58.90&17.22\\\hline
	\mr{2}{\Th{Score-CAM}}& Baseline &26.49&18.62&33.84&58.42&18.31\\
		& Ours &24.82&19.49&35.51&59.11&18.34\\\hline
	\mr{2}{\Th{Ablation-CAM}}& Baseline &31.96&14.02&28.33&58.36&17.14\\
	& Ours &29.90&15.03&29.61&58.70&17.37\\\hline
	\mr{2}{\Th{Axiom-CAM}}& Baseline &30.16&15.23&29.98&58.47&17.47\\
		& Ours &28.09&16.20&31.53&58.76&17.57\\\midrule

	\mc{7}{\Th{MobileNet-V2}}\\\midrule
	\Th{Method}&\Th{Error}&\Th{AD$\downarrow$}&\Th{AG$\uparrow$}&\Th{AI$\uparrow$}&\Th{Ins$\uparrow$}&\Th{Del$\downarrow$}\\\hline
	\mr{2}{\Th{Grad-CAM}}& Baseline &44.64&6.57&25.62&44.64&14.34\\
	& Ours &40.89&7.31&27.08&45.57&15.20\\\hline
	\mr{2}{\Th{Grad-CAM++}}& Baseline &45.98&6.12&24.10&44.72&14.76\\
		& Ours &40.76&6.85&26.46&45.51&14.92\\\hline
	\mr{2}{\Th{Score-CAM}}& Baseline &40.55&7.85&28.57&45.62&14.52\\
		& Ours &36.34&9.09&30.50&46.35&14.72\\\hline
	\mr{2}{\Th{Ablation-CAM}}& Baseline &45.15&6.38&25.32&44.62&15.03\\
		& Ours &41.13&7.03&26.10&45.38&15.12\\\hline
	\mr{2}{\Th{Axiom-CAM}}& Baseline &44.65&6.57&25.62&44.64&15.27\\
		& Ours &40.89&7.31&27.08&45.57&15.20\\\bottomrule
\end{tabular}
\caption{\emph{Interpretability metrics} of standard \vs our training using ResNet-18 and MobileNet-V2 on CIFAR-100. Using cosine error function for our training.}
\label{tab:C100_quant}
\end{table}

\subsection{Quantitative results}

We evaluate the effect of training a given model using our proposed approach with \textit{faithfulness} and \textit{causality} metrics. As shown in \autoref{tab:C100_acc} and \autoref{tab:C100_quant}, we obtain improvements on both networks and on four out of five interpretability metrics, while remaining within half percent or improving accuracy relative to the baseline, standard backpropagation.

The improvements are higher for faithfulness metrics AD, AG, and AI. Insertion gets a smaller but consistent improvement. Deletion is mostly inferior with our method, but with a very small difference. This may be due to limitations of the metrics, as reported in previous works~\cite{zhang2023opti}.

It is interesting to note that improvements on Score-CAM mean that our training not only improves gradient but also builds better activation maps, since Score-CAM only relies on those.

\begin{table}
\centering
\scriptsize
\setlength{\tabcolsep}{3.5pt}
\begin{tabular}{lcccccc} \toprule
	\Th{Error function}&\Th{Acc}&\Th{AD$\downarrow$}&\Th{AG$\uparrow$}&\Th{AI$\uparrow$}&\Th{Ins$\uparrow$}&\Th{Del$\downarrow$}\\\midrule
	Baseline  & 73.42 & 30.16      & 15.23      & 29.99      & 58.47      & 17.47 \\ \midrule
	Cosine    & 72.86 & \tb{28.09} & \tb{16.19} & \tb{31.53} & 58.76      & 17.57 \\
	Histogram & 73.88 & 30.39      & 14.78      & 29.38      & 58.52      & \textbf{17.35} \\
	MAE       & 73.41 & 30.33      & 15.06      & 29.61      & 58.13      & 17.95 \\
	MSE       & 73.86 & 29.64      & 15.19      & 30.11      & \tb{59.05} & 18.02 \\ \bottomrule
\end{tabular}
\caption{Effect of \emph{error function} on our approach, using ResNet-18 and Grad-CAM attributions on CIFAR-100.}
\label{tab:Regs}
\end{table}

\begin{table}
\centering
\scriptsize
    \begin{tabular}{lcccccc} \toprule
    $\lambda$&\Th{Acc}&\Th{AD$\downarrow$}&\Th{AG$\uparrow$}&\Th{AI$\uparrow$}&\Th{Ins$\uparrow$}&\Th{Del$\downarrow$}\\\midrule
    $0$  &73.42&30.16&15.23&29.99&58.47&17.47\\
    $1\times10^{-3}$ &\tb{73.71}&29.52&15.17&30.03&59.23&\tb{17.45}\\
    $2.5\times10^{-3}$ &72.99&30.53&15.82&30.56&59.04&17.96\\
    $5\times10^{-3}$ &72.46&30.10&16.06&30.67&57.47&17.80\\
    $7.5\times10^{-3}$ &72.86&\tb{28.09}&\tb{16.20}&\tb{31.53}&58.76&17.57\\
    $1\times10^{-2}$ &73.28&28.97&15.75&31.16&58.99&17.50\\
    $1\times10^{-1}$ &73.00&28.93&16.13&31.55&\tb{59.66}&17.95\\
    $1$ &73.30&28.44&16.02&31.31&58.64&17.48\\
    $10$ &73.04&29.28&15.23&30.47&58.74&17.47\\\bottomrule
    \end{tabular}
    \caption{Effect of \emph{regularization coefficient} $\lambda$~\eq{total} on our approach, using ResNet-18 and Grad-CAM attributions on CIFAR-100. Using cosine error function for our training.}
    \label{tab:variation}
\end{table}

\subsection{Ablation Experiments}

Using ResNet-18 and Grad-CAM attributions, we analyze the effect of the error function and the regularization coefficient $\lambda$~\eq{total} on our approach.

\paragraph{Error function}

As shown in \autoref{tab:Regs}, we obtain a consistent improvement on most metrics for all error functions. Accuracy remains stable within half percent of the original model. However, most options have little or negative effect on deletion. Cosine similarity provides improvements in most metrics, while maintaining deletion performance. We thus choose cosine error function by default.

\paragraph{Regularization coefficient}

As shown in \autoref{tab:variation}, our method is not very sensible to the regularization coefficient $\lambda$. The value of $7.5 \times 10^{-3}$ works better in general and is thus selected as default.

\uppercase{\section{conclusion}}

In this paper, we propose a new training approach to improve the gradient of a CNN in terms of interpretability. Our method forces the gradient with respect to the input image obtained by backpropagation to align with the gradient coming from guided backpropagation. The results of our training are evaluated according to several interpretability methods and metrics. Our method offers consistent improvement on most metrics for two networks, while remaining within a small margin of the standard gradient in terms accuracy.

\uppercase{\section{Acknowledgement}}
This work has received funding from the UnLIR ANR project (ANR-19-CE23-0009), the Excellence Initiative of Aix-Marseille Universite - A*Midex, a French “Investissements d’Avenir programme” (AMX-21-IET-017). Part of this work was performed using HPC resources from GENCI-IDRIS (Grant 2020-AD011011853 and 2020-AD011013110).


\bibliographystyle{apalike}
\footnotesize
\bibliography{egbib}

\end{document}